\documentclass[letterpaper]{article} 
\usepackage{aaai25}  
\usepackage{times}  
\usepackage{helvet}  
\usepackage{courier}  
\usepackage[hyphens]{url}  
\usepackage{graphicx} 
\usepackage{natbib}  
\usepackage{caption} 
\usepackage{algorithm}
\usepackage{algorithmic}

\usepackage{newfloat}
\usepackage{listings}
\DeclareCaptionStyle{ruled}{labelfont=normalfont,labelsep=colon,strut=off} 
\floatstyle{ruled}
\newfloat{listing}{tb}{lst}{}
\floatname{listing}{Listing}
\usepackage{amsmath}
\usepackage{booktabs}
\usepackage{xcolor}
\usepackage{mdframed}
\usepackage{subcaption}
\usepackage{multirow}
\usepackage{xspace}
\usepackage[utf8]{inputenc}

\newcommand{\UserSumBench}{\textsc{UserSumBench}\xspace}

\nocopyright

\title{\UserSumBench: A Benchmark Framework for Evaluating User Summarization Approaches}
\author {
    Chao Wang\textsuperscript{\rm 1},
    Neo Wu\textsuperscript{\rm 1},
    Lin Ning\textsuperscript{\rm 1},
    Jiaxing Wu\textsuperscript{\rm 1},
    Luyang Liu\textsuperscript{\rm 1},
    Jun Xie\textsuperscript{\rm 1},
    Shawn O'Banion\textsuperscript{\rm 1},
    Bradley Green\textsuperscript{\rm 1}
}
\affiliations {
    \textsuperscript{\rm 1}Google DeepMind
}

\usepackage{bibentry}

\begin{document}

\maketitle

\begin{abstract}
Large language models (LLMs) have shown remarkable capabilities in generating user summaries from a long list of raw user activity data. These summaries capture essential user information such as preferences and interests, and therefore are invaluable for LLM-based personalization applications, such as explainable recommender systems.
However, the development of new summarization techniques is hindered by the lack of ground-truth labels, the inherent subjectivity of user summaries, and human evaluation which is often costly and time-consuming.
To address these challenges, we introduce \UserSumBench, a benchmark framework designed to facilitate iterative development of LLM-based summarization approaches. This framework offers two key components: (1) A reference-free summary quality metric. We show that this metric is effective and aligned with human preferences across three diverse datasets (MovieLens, Yelp and Amazon Review). (2) A novel robust summarization method that leverages time-hierarchical summarizer and self-critique verifier to produce high-quality summaries while eliminating hallucination. This method serves as a strong baseline for further innovation in summarization techniques.

\end{abstract}

\section{Introduction}

\begin{figure}[ht]
  \centering
  \includegraphics[width=0.47\textwidth]{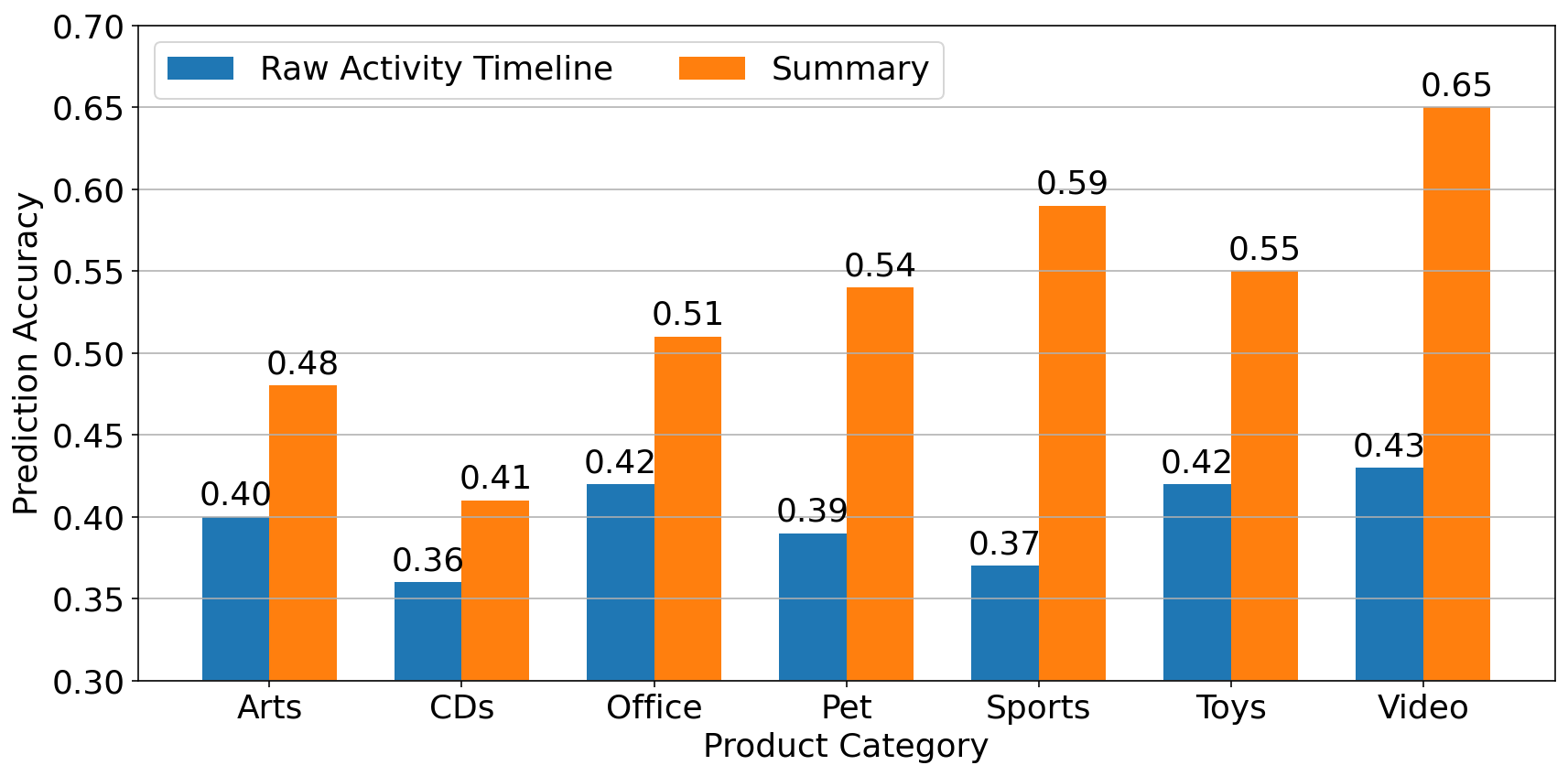}
  \caption{Comparison of next product prediction accuracy across different contexts in the Amazon Review dataset, contrasting performance using raw timelines versus summarized data.}
  \label{figure:comparison_different_context}
\end{figure}

User activity timelines, including data such as place visit histories, product reviews, movie ratings, and other digital interactions, offer valuable insights into individual preferences, behaviors, and evolving interests. These timelines are crucial for applications like personalized recommendations and user behavior analysis \cite{wang2019sequential}. Summarizing these timelines into concise, actionable insights is essential for enhancing recommendation systems and understanding user engagement trends. For example, as shown in Figure~\ref{figure:comparison_different_context}, the next product prediction accuracy of an LLM-based model on the Amazon Review dataset \cite{ni2019justifying} significantly improves when using summaries instead of raw activity timelines.

However, generating high-quality user summaries is challenging due to the complexity and diversity of user timelines. The subjective nature of summary evaluation and the lack of standardized ground-truth datasets further complicate the process. Current methods often rely on simplistic heuristics or models that struggle with these issues \cite{giarelis2023abstractive}. Moreover, the absence of standardized benchmarks or reliable metrics hampers the evaluation of summarization effectiveness \cite{fabbri2021summeval,lloret2018challenging}.

To tackle these challenges, we introduce \UserSumBench, a comprehensive benchmark framework specifically designed to evaluate user summarization approaches by assessing the quality of user summaries generated from activity timelines. \UserSumBench consists of two key components: a robust, reference-free summary quality metric and a strong baseline summarization approach.

\begin{figure*}[t]
  \centering
  \includegraphics[width=0.8\textwidth]{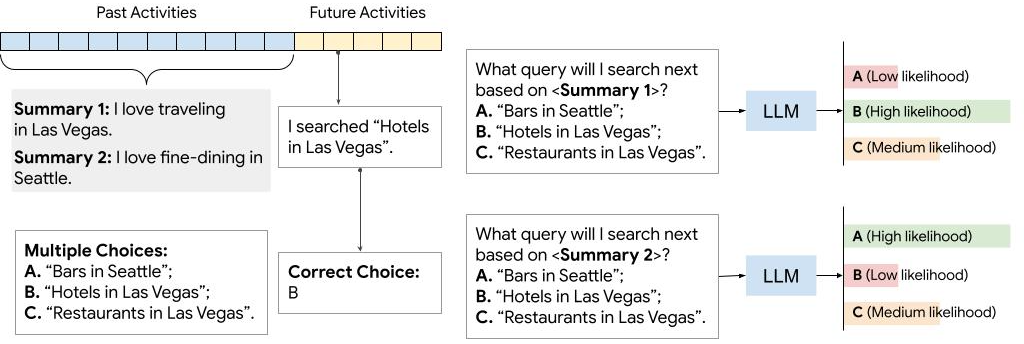}
  \caption{Evaluating summary quality through future activity prediction tasks, where LLMs predict the most likely user queries based on generated summaries of past activities.}
  \label{figure:sumamry_eval_likelihood}
\end{figure*}

In \UserSumBench, the proposed quality metric evaluates the effectiveness of user summarization approaches by measuring how accurately the generated summaries predict future user activities. This metric offers a quantitative assessment of how well the summaries capture key aspects of user behavior (see Figure~\ref{figure:sumamry_eval_likelihood}) and has demonstrated strong alignment with human ratings.

The proposed strong baseline summarization approach employs a time-hierarchical and self-critique method. This approach uses an LLM for initial summarization, followed by iterative refinement to reduce hallucinations and improve summary quality. This baseline not only validates the benchmark metrics but also serves as a foundation for future innovations in summarization techniques.

\textbf{Key Contributions:}
\begin{itemize}
\item Introduction of a quality metric for evaluating user summarization approaches through user summaries, demonstrating strong alignment with human ratings, thereby validating its effectiveness and simplifying the evaluation process.
\item Introduction of a strong baseline summarization approach, a time-hierarchical and self-critique method, setting the foundation for future advancements in summarization techniques.
\end{itemize}

\section{Related Works}

The lack of standardized benchmarks has long been a challenge in evaluating summarization approaches \cite{fabbri2021summeval}. While datasets like MovieLens \cite{harper2015movielens} and Amazon Reviews \cite{ni2019justifying} offer comprehensive logs of user activities, they lack corresponding ground-truth summaries, complicating the assessment of summarization techniques. Efforts have been made to create datasets that pair user activities with manually crafted summaries. For example, Liu et al. \cite{liu2023benchmarking} emphasized the importance of performance prediction in summarization evaluation, advocating for benchmarks that measure the predictive power of generated summaries. Similarly, Akkasi et al. \cite{akkasi2023reference} explored reference-free evaluation methods, proposing metrics designed to assess summaries' ability to convey essential content relevant to future activities. These studies highlight the limitations of current evaluation practices and the need for more predictive and reliable benchmarks.

Recent research has focused on establishing standardized evaluation methodologies for summarization \cite{fabbri2021summeval,liu2023benchmarking,chen2024rethinking}. Fabbri et al. \cite{fabbri2021summeval} addressed the shortcomings in existing summarization evaluation methods by reassessing 14 automatic evaluation metrics using outputs from recent neural summarization models. They also released a toolkit of evaluation metrics to promote consistency in reporting results. Chen et al. \cite{chen2024rethinking} proposed a facet-aware evaluation paradigm for scientific abstracts, introducing benchmarks that enable more nuanced comparisons of evaluation metrics in this context.

\section{\UserSumBench Framework}

In this section, we present the two key components of the \UserSumBench framework: Benchmark Metrics and Hierarchy-Critique Summary Generation. The Benchmark Metrics assess summarization approaches using various criteria, such as future user activity predictions, to ensure a comprehensive evaluation of the generated summaries. The Hierarchy-Critique Summary Generation method provides a strong baseline for evaluating and improving these techniques.

\subsection{Benchmark Metrics}
\label{sec:benchmark_metrics}

\UserSumBench includes three evaluation metrics to assess different aspects of user summarization approaches.

\subsubsection{Quality Metric}
\label{sec:quality_metric}

Quality Metric is designed to evaluate the predictive accuracy of user summaries in forecasting future activities. The evaluation involves splitting each user activity timeline into past and future activities (see Figure~\ref{figure:sumamry_eval_likelihood}). Summaries are generated from the past activities, and the quality of these summaries is measured by how well they predict the future activities. The quality of a user summary, \( Q_s \), is computed by aggregating performance across multiple future activity prediction tasks.

\begin{align}
Q_s = I \left[ \sum_{t}^{T_s} q_{s,t} \geq m \right]
\label{eq:qs}
\end{align}

Here, \( q_{s,t} \) denotes the binary prediction outcome for summary \( s \) on task \( t \) from the set of tasks \( T_s \); \( m \) is the threshold for the number of correct predictions required; \( I[.] \) is the indicator function, where a result of 1 indicates a "Good" summary, and 0 indicates a "Bad" summary.

To evaluate a summarization approach on a generated summary set \( S \), the Quality Metric (QM) is calculated as the percentage of summaries classified as "Good" based on the qualities of these summaries.

\begin{align}
QM = \frac{\sum_{s \in S} Q_s}{|S|}
\label{eq:qm}
\end{align}

\subsubsection{Instruction Following Metric}

The Instruction Following Metric evaluates how well the user summaries adhere to specific constraints, such as a word limit, as introduced in other works \cite{skopek2023towards}. For a summarization approach, the Instruction Following Metric (IFM) is defined as the proportion of summaries within the set \( S \) that meet the word limit constraint \( X \).

\begin{align}
IFM = \frac{|\{ s \in S \mid \text{length}(s) \leq X \}|}{|S|}
\label{eq:ifm}
\end{align}

While this metric was originally proposed in earlier studies, it remains a useful measure for assessing how effectively a summarization approach adheres to the given prompt instructions.

\subsubsection{Information Density Metric}

The Information Density Metric assesses the conciseness and informativeness of user summaries, combining elements of both the Quality Metric and the Instruction Following Metric. This metric evaluates the balance between the length of a summary and its effectiveness in predicting future activities. For each summary \( s \) in the set \( S \), and each associated task set \( T_s \), the Information Density Metric (IDM) is calculated by dividing the average task prediction accuracy by the length of the summary.

\begin{align}
IDM = \frac{1}{|S|} \sum_{s \in S} \left( \frac{\frac{1}{|T_s|} \sum_{t \in T_s} \text{acc}(t)}{\text{length}(s)} \right)
\label{eq:idm}
\end{align}

This metric provides a quantitative approach to evaluate the trade-off between informativeness and brevity in user summaries, ensuring that summaries are both concise and meaningful.

\subsection{Hierarchy-Critique Summary Generation}

\begin{figure*}[ht]
  \centering
  \includegraphics[width=0.9\textwidth]{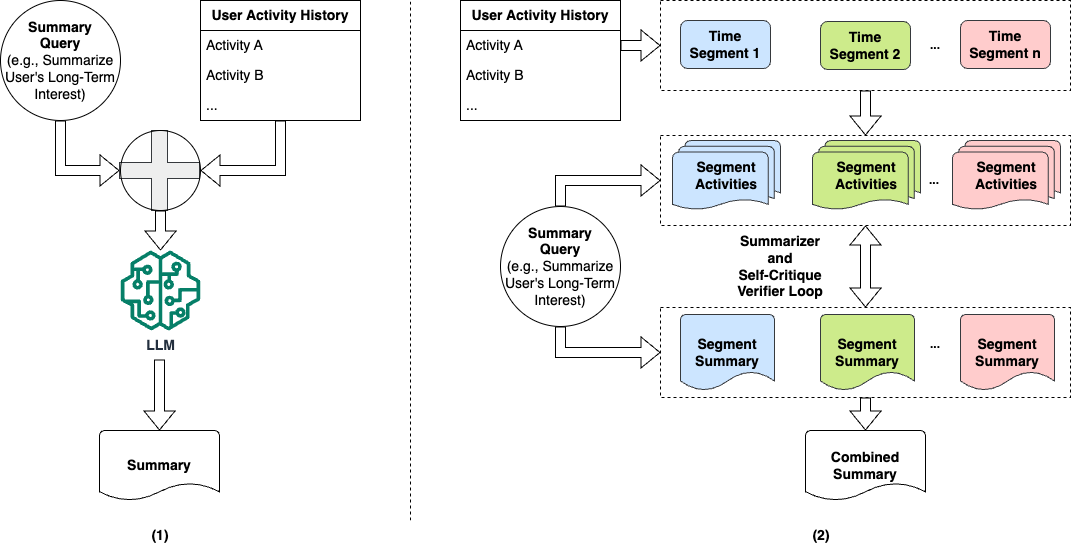}
  \caption{Comparison of summarization approaches: (1) Single-step summarization approach, where a summary is generated directly from the user's activity history; (2) Time-hierarchical and self-critique summarization approach, which involves segmenting the user's activity history over time, summarizing each segment, and iteratively refining the summaries before combining them into a final summary.}
  \label{figure:summarization_approaches}
\end{figure*}

\UserSumBench introduces a Time-Hierarchical and Self-Critique (Hierarchy-Critique) summarization approach (see Figure~\ref{figure:summarization_approaches}(2)), which is demonstrated to outperform a simpler single-step method (see Figure~\ref{figure:summarization_approaches}(1), refer to Appendix~\ref{appendix:single_step_summarization_prompt} for the prompt details). For more details on this comparison, please refer to Section~\ref{sec:eval_approaches}. The Hierarchy-Critique approach is designed to address the challenges of generating factually consistent summaries by mitigating hallucinations while maintaining computational efficiency through a time-hierarchical structure.

The Hierarchy-Critique approach works by first segmenting a user's activity history into manageable time intervals, ensuring each segment meets a minimum activity threshold to provide a comprehensive representation of the user's behavior. This segmentation allows LLMs to process the data efficiently within their context window limits. As depicted in Figure~\ref{figure:iterative_summarization_refinement}, the summarizer model generates an initial summary for each segment (refer to Appendix~\ref{appendix:segment_summarization_prompt}). These segment summaries are then refined by a verifier model \cite{wang2023self}, whose role is to identify and correct potential hallucinations (e.g., query inconsistencies, factual inaccuracies), ensuring that each segment accurately reflects the user's activities. Finally, the refined segment summaries are synthesized into a cohesive summary that encapsulates the user's overall behavior and preferences (refer to Appendix~\ref{appendix:combine_segments}), with all time segments combined in chronological order.

\begin{algorithm}[tb]
\caption{Hierarchy-Critique Summary Generation}
\label{alg:time_hierarchical_self_critique}
\textbf{Input}: User activity history \( \Phi \)\\
\textbf{Parameter}: Time segments \( \{T_1, T_2, \ldots, T_n\} \), Summarizer model \( LLM_{sum} \), Verifier model \( LLM_{ver} \)\\
\textbf{Output}: Factually consistent user summary \( S \)

\begin{algorithmic}[1]
\STATE Divide user activity history \( \Phi \) into time segments \( \{T_1, T_2, \ldots, T_n\} \)
\FOR{each time segment \( T_i \)}
    \STATE Generate initial summary \( S_i \) for segment \( T_i \) using \( LLM_{sum} \)
    \STATE Refine \( S_i \) using \( LLM_{ver} \):
        \STATE \quad Identify and correct \textbf{Query Inconsistency}
        \STATE \quad Identify and correct \textbf{Fact Inconsistency}:
            \STATE \quad \quad Extract key KG entities \( \{e_{i,1}, e_{i,2}, \ldots, e_{i,k}\} \) from \( S_i \)
            \STATE \quad \quad Generate question-answer pairs \( \{(q_{i,1}, a_{i,1}), (q_{i,2}, a_{i,2}), \ldots, (q_{i,k}, a_{i,k})\} \) using \( LLM_{ver} \) on \( S_i \)
            \STATE \quad \quad Verify consistency of each pair \( (q_{i,j}, a_{i,j}) \) with user activities \( \Phi \)
            \STATE \quad \quad For inconsistent pairs, regenerate \( S_i \) incorporating feedback
\ENDFOR
\STATE Synthesize segment summaries \( \{S_1, S_2, \ldots, S_n\} \) into a single cohesive summary \( S \)
\STATE \textbf{return} \( S \)
\end{algorithmic}
\end{algorithm}

As illustrated in Figure~\ref{figure:iterative_summarization_refinement}, the Hierarchy-Critique approach employs two LLMs (a summarizer and a verifier) to iteratively refine segment summaries. These LLMs can be the same model or different models, with the verifier focusing on identifying specific types of hallucinations:

\begin{figure*}[t]
  \centering
  \includegraphics[width=0.8\textwidth]{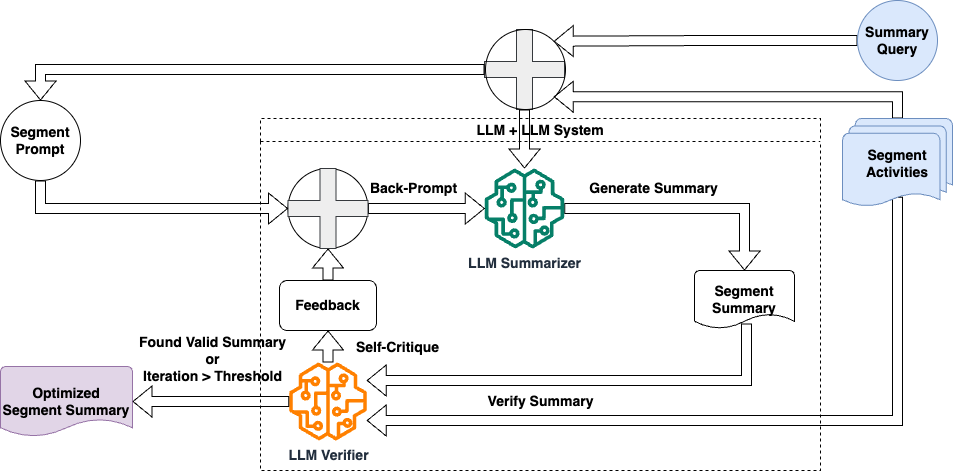}
  \caption{Diagram of the iterative summarization refinement process, where the LLM Summarizer generates an initial segment summary, which is then iteratively critiqued and refined by the LLM Verifier until an optimized summary is achieved or a specified iteration threshold is met.}
  \label{figure:iterative_summarization_refinement}
\end{figure*}

\begin{itemize}
    \item \textbf{Query Consistency:} Ensuring that the summary is relevant to the initial query. The verifier checks for consistency between the query and the summary based on a provided prompt (refer to Appendix~\ref{appendix:query_consistency_prompt}).

    \item \textbf{Fact Consistency:} Ensuring that the information in the generated summary is accurate and consistent with the user's activities. To identify factual inconsistencies, we propose using the Question Generation - Question Answering (QG-QA) method \cite{xu2024identifying}, which operates on key Knowledge Graph (KG) entities \cite{singhal2012kg} extracted from the summary. This method involves two components: Question Generation and Question Answering.

    \begin{itemize}
        \item \textbf{Question Generation (QG):} Given a summary and an KG entity (e.g., Hiking /m/012v4j) extracted from the summary, the verifier generates a question-answer pair based on the summary context using a user-specified prompt (refer to Appendix~\ref{appendix:qg_prompt}). The answer is the KG entity.
        \item \textbf{Question Answering (QA):} The consistency of these question-answer pairs is then verified against the user's activities using the verifier, guided by a user-specified prompt (refer to Appendix~\ref{appendix:qa_prompt}).
    \end{itemize}

    Let \( S \) be the summary candidate based on the retrieved activities \( \Phi \), \( e_k \) the \( k \)-th KG entity in \( S \), \( (q_k, a_k) \) its corresponding question-answer pair, and \( h_k \) the binary result (\texttt{consistent}, \texttt{inconsistent}) of \( (q_k, a_k) \) compared to \( \Phi \). The QG-QA process can be represented by Equations~\ref{eq:qg} and ~\ref{eq:qa}.
    
    \begin{align}
    \textbf{QG:} & \quad \left\{ \left( q_k, a_k \right) \right\} = LLM_{QG} \left( S, \left\{ e_k \right\} \right)
    \label{eq:qg}
    \end{align}
    
    \begin{align}
    \textbf{QA:} & \quad \left\{ h_k \right\} = LLM_{QA} \left( \Phi, \left\{ \left( q_k, a_k \right) \right\} \right)
    \label{eq:qa}
    \end{align}
    
    For any inconsistent question-answer pairs identified by \( h_k \), the summary is regenerated by incorporating feedback from these hallucinations into the original prompt (refer to Appendix~\ref{appendix:segment_summarization_prompt_with_feedback}), as described in Figure~\ref{figure:iterative_summarization_refinement} and detailed in Algorithm~\ref{alg:time_hierarchical_self_critique}.
\end{itemize}

\section{Evaluation}

In this section, we validate the benchmark metrics and evaluate hierarchy-critique summarization approach within the \UserSumBench framework.

\subsection{Validating Benchmark Metrics}

To validate the \UserSumBench benchmark metrics, we study their alignment with human ratings on three public user activity datasets: MovieLens 1M \cite{harper2015movielens}, Yelp \cite{yelp2023dataset}, and Amazon Review \cite{ni2019justifying}.

\subsubsection{Datasets and Evaluation Tasks}
This section outlines the dataset preparation and the prediction tasks used for evaluation.

User timelines from the three datasets were filtered based on activity count \( L_{\Phi} \) to ensure a balance between data sufficiency and computational efficiency:

\begin{itemize}
    \item \( L_{\Phi} \ge N_{low} \): Timelines with fewer than 50 activities were excluded to ensure sufficient context for generating robust summaries.
    \item \( L_{\Phi} \le N_{up} \): Timelines with more than 200 activities were truncated to the most recent 200 to focus on recent behavior patterns and manage computational load.
\end{itemize}

These thresholds (\( N_{low} = 50 \) and \( N_{up} = 200 \)) were chosen to balance informative summaries with processing efficiency. Table \ref{table:number_of_examples} shows the number of examples used for evaluation after filtering.

\begin{table}[ht]
  \centering
  \begin{tabular}{c|c}
  \toprule
  \textbf{Dataset} & \textbf{Number of Examples} \\
  \midrule
  \textbf{\small MovieLens} & 4297 \\
  \textbf{\small Yelp} & 6798 \\
  \textbf{\small Amazon Review} & 5046 \\
  \bottomrule
  \end{tabular}
  \vskip 0.1in
  \caption{Number of examples used for evaluating summarization metrics across different datasets.}
  \label{table:number_of_examples}
\end{table}

Four prediction tasks were defined to evaluate the quality of the summaries generated from these datasets. Each task was structured as a multiple-choice question with one correct answer and four randomly selected incorrect options. The order of all choices was randomized during execution to ensure robustness.

\begin{itemize}
    \item \( t_1 \): Predict the user's next activity (e.g., next watched movie name, next purchased product name) based solely on the summary, assessing the summary’s ability to encapsulate immediate user behavior.
    \item \( t_2 \): Predict the user's next activity considering both the summary and the \( N_r \) most recent activities, evaluating the summary’s effectiveness in conjunction with recent user data.
    \item \( t_3 \): Predict the category of the user's next activity (e.g., next watched movie genre, next purchased product category) based on the summary alone, testing the summary’s ability to generalize user preferences.
    \item \( t_4 \): Predict the category of the user's next activity using both the summary and the \( N_r \) most recent activities, examining how well the summary integrates with recent behavior to provide accurate insights.
\end{itemize}

These tasks were designed to comprehensively assess different aspects of summary quality. The default value for \( N_r = 20 \) and the threshold \( m = 3 \) as defined in Equation~\ref{eq:qs}.

\subsubsection{Quality Metric vs. Human Ratings}

To validate the effectiveness of the proposed Quality Metric (refer to Section~\ref{sec:quality_metric}), we conducted a human rating exercise.

We selected 200 examples from each of the three summary-extended datasets (MovieLens, Yelp, and Amazon Reviews), resulting in a total of 600 examples. These examples were based on user activity histories from English-speaking residents in the United States, and the summaries were generated using the single-step summarization approach (see Figure~\ref{figure:summarization_approaches}(1)) with the Gemini Advanced model \cite{team2023gemini}.

Six human raters (three male and three female), all English-speaking USA residents familiar with common user activities participated in the evaluation. Each dataset was evaluated by two raters (one male and one female) to enhance robustness and minimize bias, with each rater assessing 100 examples.

Given the subjective nature of summary evaluation, particularly for criteria like "Fact Hallucinations Verification," a unified rating guideline was implemented to standardize the process across all raters. This guideline was designed to ensure consistency and reliability in the ratings, classifying summaries as either "Good" or "Bad" based on the following criteria:

\begin{itemize}

\item \textbf{Query Hallucinations Verification:} Assesses whether the summary accurately reflects the specified summary query. For instance, if the query is to "summarize a user’s movie-watching preferences," but the summary only lists movie-watching activities without indicating any preferences, the summary would fail this check.

\item \textbf{Fact Hallucinations Verification:} Evaluates whether the details in the summary align with the context of the input activities. For example, if a user's recent activity history shows a focus on purchasing toys, but the summary inaccurately states that the user mainly purchased office supplies, it would be marked as inaccurate.

\item \textbf{Top Category Recall:} Checks whether at least one of the top three categories, such as movie genres, is mentioned in the summary. Very popular categories like "Restaurant" in the Yelp dataset are excluded from the top three to ensure meaningful assessment.

\end{itemize}

This structured approach, with clearly defined evaluation criteria and qualified raters, ensures that the human annotation process is consistent and capable of producing reliable comparisons between prediction results and human ratings.

After obtaining prediction results and human ratings for the same user summaries, we categorized each summary into one of four possible outcomes based on the alignment between the prediction and the human rating. We then calculated the \textbf{Metric-Annotator Agreement (MAA)} using Equation~\ref{eq:mma}, which represents the percentage of cases where the evaluation metrics and human annotators agreed on the quality of the summary. Higher scores indicate better alignment.

\begin{align}
MAA = \frac{TP + TN}{TP + FP + FN + TN}
\label{eq:mma}
\end{align}

In this equation, the four possible outcomes are as follows:

\begin{itemize}
    \item \( TP \) (True Positive): Both the prediction and the human rating classify the summary as "Good."
    \item \( FP \) (False Positive): The prediction classifies the summary as "Good," but the human rating classifies it as "Bad."
    \item \( FN \) (False Negative): The prediction classifies the summary as "Bad," but the human rating classifies it as "Good."
    \item \( TN \) (True Negative): Both the prediction and the human rating classify the summary as "Bad."
\end{itemize}

As shown in Table~\ref{table:mma_metrics_ratings}, the proposed Quality Metric demonstrated strong alignment with human ratings, with above 70\% agreement across all datasets.

\begin{table}
  \centering
  \begin{tabular}{c|c}
  \toprule
  \textbf{Dataset} & \textbf{MAA} \\
  \midrule
  \text{MovieLens} & 71.0\% \\
  \text{Yelp} & 74.0\% \\
  \text{Amazon Review} & 73.0\% \\
  \bottomrule
  \end{tabular}
  \vskip 0.1in
  \caption{Metric-Annotator Agreement (MAA) between Quality Metric and human ratings across different datasets.}
  \label{table:mma_metrics_ratings}
\end{table}

To further evaluate the performance of the proposed benchmark metrics, we applied them to three popular models: Gemini 1.5 Pro \cite{reid2024gemini}, GPT-4o \cite{achiam2023gpt}, and Claude 3 Haiku \cite{anthropic2023claude3}. Detailed comparisons can be found in Appendix~\ref{appendix:benchmark_comparisons_models}.

\begin{table}[ht]
  \centering
  \begin{tabular}{l|ccc}
  \toprule
  \textbf{Metric} & \textbf{MovieLens} & \textbf{Yelp} & \textbf{Amazon} \\
  \midrule
  \textbf{ROUGE-2} & 0.163 & 0.037 & 0.289 \\
  \textbf{ROUGE-L} & 0.151 & 0.147 & 0.338 \\
  \textbf{BLEU} & 0.049 & 0.043 & -0.077 \\
  \textbf{BERTScore-precision} & 0.077 & -0.021 & 0.254 \\
  \textbf{BERTScore-recall} & -0.006 & 0.090 & 0.146 \\
  \textbf{BERTScore-F1} & -0.119 & 0.071 & -0.127 \\
  \textbf{BLEURT} & -0.082 & 0.088 & -0.023 \\
  \textbf{AutoEval} & 0.204 & 0.120 & \textbf{0.490} \\
  \midrule
  \textbf{Quality $Q_s$} & \textbf{0.363} & \textbf{0.458} & 0.366 \\
  \bottomrule
  \end{tabular}
  \vskip 0.1in
  \caption{Comparison of different evaluation metrics and summary quality measurement $Q_s$ across the MovieLens, Yelp, and Amazon Review datasets.}
  \label{table:evaluation_metrics_comparison}
\end{table}

\begin{table*}[ht]
  \centering
  \begin{tabular}{l|c|cc|c}
  \toprule
  \textbf{Metric} & \textbf{Dataset} & \textbf{Single-Step} & \textbf{Hierarchy-Critique} & \textbf{Increment Percentage} \\
  \midrule
  \multirow{3}{*}{\textbf{Quality Metric}} 
  & MovieLens       & 0.557 $\pm$ 0.002 & \textbf{0.586 $\pm$ 0.003} & \textcolor{blue}{5.21\%} \\
  & Yelp            & 0.459 $\pm$ 0.002 & \textbf{0.512 $\pm$ 0.003} & \textcolor{blue}{11.55\%} \\
  & Amazon Review   & 0.616 $\pm$ 0.002 & \textbf{0.631 $\pm$ 0.002} & \textcolor{blue}{2.44\%} \\
  \midrule
  \multirow{3}{*}{\textbf{Instruction Following Metric}} 
  & MovieLens       & 0.836 & \textbf{0.988} & \textcolor{blue}{18.18\%} \\
  & Yelp            & 0.791 & \textbf{0.992} & \textcolor{blue}{25.41\%} \\
  & Amazon Review   & 0.842 & \textbf{0.999} & \textcolor{blue}{18.65\%} \\
  \midrule
  \multirow{3}{*}{\textbf{Information Density Metric (x0.1\%)}} 
  & MovieLens       & 3.788 $\pm$ 0.004 & \textbf{5.184 $\pm$ 0.005} & \textcolor{blue}{36.85\%} \\
  & Yelp            & 3.561 $\pm$ 0.004 & \textbf{4.721 $\pm$ 0.005} & \textcolor{blue}{32.58\%} \\
  & Amazon Review   & 2.145 $\pm$ 0.002 & \textbf{3.202 $\pm$ 0.003} & \textcolor{blue}{49.28\%} \\
  \bottomrule
  \end{tabular}
  \vskip 0.1in
  \caption{Comparison of quality metric, instruction following metric, and information density metric between Single-Step and Hierarchy-Critique approaches across different datasets.}
  \label{table:combined_metrics_comparison}
\end{table*}

In addition to \UserSumBench benchmark metrics, we also considered other reference-free evaluation metrics that do not require ground-truth summaries for user summary evaluation. These metrics compare the generated summary against user activities without relying on a reference summary.

\begin{itemize}
    \item \textbf{ROUGE-2} \cite{lin2004rouge}: Measures the overlap of bigrams between the generated summary and user activities.
    \item \textbf{ROUGE-L} \cite{lin2004rouge}: Measures the Longest Common Subsequence (LCS) between the generated summary and user activities.
    \item \textbf{BLEU} \cite{papineni2002bleu}: Evaluates the precision of n-grams in the generated summary compared to user activities.
    \item \textbf{BertScore} \cite{zhang2019bertscore}: Uses BERT embeddings to compute precision, recall, and F1 score based on the similarity of words in the generated summary and user activities.
        \begin{itemize}
            \item \textbf{BertScore-precision:} Measures the precision of embedding overlap.
            \item \textbf{BertScore-recall:} Measures the recall of embedding overlap.
            \item \textbf{BertScore-F1:} Combines precision and recall to provide an F1 score.
        \end{itemize}
    \item \textbf{BLEURT} \cite{sellam2020bleurt}: Uses a pre-trained model to evaluate the quality of the generated text based on human judgments compared to user activities.
    \item \textbf{AutoEval} \cite{chiang2023closer}: Adopts an LLM to automatically evaluates summaries without requiring reference summaries, focusing on coherence and relevance.
\end{itemize}

To further validate our approach, we calculated the Pearson Correlation Coefficient between human ratings and the reference-free metrics, including the proposed summary quality measurement $Q_s$ (see Equation~\ref{eq:qs}). In this comparison, BLEURT utilized a pre-trained BLEURT-base-128 model \cite{sellam2020bleurt} for evaluation, while AutoEval was based on the Gemini 1.5 Pro model. The $Q_s$ also employed the Gemini 1.5 Pro model for future activity predictions. As shown in Table~\ref{table:evaluation_metrics_comparison}, our metric demonstrated a more reliable and higher correlation with human ratings than other metrics, as they are specifically tailored to the characteristics of user timeline activities. Despite not having a perfect correlation, the $Q_s$ remains highly useful for tasks like weakly supervised learning \cite{zhou2018brief}, where ground-truth summaries are limited.

\subsection{Evaluating Summarization Approaches}
\label{sec:eval_approaches}

In this section, we evaluate our proposed Hierarchy-Critique summarization approach using the \UserSumBench benchmark metrics. For consistency, all LLMs in this experiment used the same model: Gemini 1.5 Flash \cite{reid2024gemini}, across summarization, self-critique verification, and prediction tasks. To validate the robustness of the benchmarks, we conducted three prediction runs on all three datasets to calculate the $Mean \pm Sd$ values for the Quality Metric and the Information Density Metric. The results, presented in Table~\ref{table:combined_metrics_comparison}, demonstrate that the Hierarchy-Critique approach consistently outperforms the Single-Step approach across various evaluation metrics. Additionally, the stability of the benchmark metrics across multiple predictions confirms their robustness.

\begin{itemize}
    \item \textbf{Quality Metric:} The Hierarchy-Critique approach achieved superior scores on the MovieLens, Yelp, and Amazon Review datasets, with increases of 5.21\%, 11.55\%, and 2.44\% respectively. These gains indicate that the iterative refinement process effectively reduces hallucinations and enhances the overall consistency and quality of the summaries.
    
    \item \textbf{Instruction Following Metric:} Significant improvements were observed with the Hierarchy-Critique approach, particularly in the Yelp dataset, which showed a 25.41\% increase, followed by MovieLens and Amazon Review with gains of 18.18\% and 18.65\% respectively. This suggests that the Hierarchy-Critique method is more adept at adhering to prompt constraints, likely due to its effective segmentation and summarization of user activities.
    
    \item \textbf{Information Density Metric:} The Hierarchy-Critique approach demonstrated substantial gains in this metric, with increases of 36.85\%, 32.58\%, and 49.28\% for the MovieLens, Yelp, and Amazon Review datasets respectively. This shows that the approach not only produces accurate summaries but also ensures that these summaries are concise and rich in information, effectively balancing brevity with informativeness.
\end{itemize}

These findings underscore the superiority of the Hierarchy-Critique approach over the Single-Step method, particularly in generating higher-quality, instruction-compliant, and information-dense summaries. The effectiveness of time segmentation, combined with iterative refinement and verification processes, plays a pivotal role in these improvements, establishing the Hierarchy-Critique approach as a more robust and reliable option for summarizing user activity timelines.

\section{Conclusion and Future Work}

In this paper, we introduced \UserSumBench, a comprehensive benchmark framework specifically designed to evaluate user summarization approaches through the assessment of user summaries generated from activity timelines. Our key contributions include (1) a reference-free quality metric for assessing user summarization approaches through summaries based on future user activity predictions, which has demonstrated strong effectiveness and close alignment with human preferences across three diverse datasets (MovieLens, Yelp, and Amazon Review), and (2) a robust summarization baseline method that combines a time-hierarchical summarizer with a self-critique verifier, yielding high-quality summaries while effectively minimizing hallucinations.

The strong alignment between our proposed quality metric and human ratings establishes \UserSumBench as a reliable and efficient tool for automated evaluation of user summarization approaches. By offering a cost-effective solution, \UserSumBench addresses the pressing need for standardized evaluation methods in the field of user summarization.

Looking ahead, we plan to expand \UserSumBench by integrating real-time summarization techniques and exploring its applicability across additional domains beyond the current datasets. These enhancements will further broaden the utility and impact of the framework. Additionally, we aim to encourage the broader research community to adopt \UserSumBench, fostering its potential to standardize user summary evaluation practices and drive innovation in the development of more accurate and robust summarization techniques. Ultimately, we believe \UserSumBench will play a significant role in advancing personalization, recommendation systems, and user understanding in the digital landscape.

\bibliography{aaai25}

\newpage
\onecolumn
\appendix

\section{Prompt Examples}

\subsection{LLM Prompt Template for Single-Step Summarization} 
\label{appendix:single_step_summarization_prompt}

\begin{mdframed}
\small {
----- Instructions -----

Summarize my ``User Activities'' and provide insights that address the query: ``\textbf{\{query\}}''. Adhere to the instructions below.

1. The summary should have the format with **Summary** and **Insights**.

2. The summary should take into account the changes of my long-term interest over time.

3. My activities are cataloged in the following ``User Activities'' section, with each separated by a newline.

4. Limit the summary to no more than \textbf{\{max\_words\}} words.
\newline
\newline
----- User Activities -----
\newline
\textbf{\{user\_activities\}}
}
\end{mdframed}

Here we define:
    
\textbf{\{query\}:} Query intention of the summarization. For example, "Summarize my long-term movie watching preference".
    
\textbf{\{max\_words\}:} Max number of words for the summary.
    
\textbf{\{user\_activities\}:} List of user activities.

\subsection{LLM Prompt Template for Segment Summarization} 
\label{appendix:segment_summarization_prompt}

\begin{mdframed}
\small {
----- Instructions -----

Summarize my ``User Activities'' related to the specified time range ``\textbf{\{time\_range\}}'' and provide insights that address the query: ``\textbf{\{query\}}''. Adhere to the instructions below.

1. The summary should have the format with **Summary** and **Insights**.

2. The summary should take into account the changes of my long-term interest over time.

3. My activities related to this time range are cataloged in the following ``User Activities'' section, with each separated by a newline.

4. Limit the summary to no more than \textbf{\{max\_words\}} words.
\newline
\newline
----- User Activities -----
\newline
\textbf{\{user\_activities\}}
}
\end{mdframed}

Here we define:

\textbf{\{time\_range\}:} Time range of the segment user activities.
    
\textbf{\{query\}:} Query intention of the summarization. For example, "Summarize my long-term movie watching preference".
    
\textbf{\{max\_words\}:} Max number of words for the summary.
    
\textbf{\{user\_activities\}:} List of user activities of the segment.

\subsection{LLM Prompt Template for Segment Summarization with Feedback} 
\label{appendix:segment_summarization_prompt_with_feedback}

\begin{mdframed}
\small {
----- Instructions -----

Summarize my ``User Activities'' related to the specified time range ``\textbf{\{time\_range\}}'', revise the below ``Previous Summary'' to be consistent with every ``Question'' and its corresponding ``ReferenceAnswer'' in the below ``Previous Question Answers''. Adhere to the instructions below.

1. For each ``Question'' in ``Previous Question Answers'', the ``Answer'' is derived from the ``Previous Summary'', while the ``ReferenceAnswer'' is based on my ``User Activities''.

2. Modify the ``Previous Summary'' to incorporate the ``ReferenceAnswer'' rather than the ``Answer'' for each ``Question'' in the ``Previous Question Answers''.

3. Ensure the new summary provides insights that address the query: ``\textbf{\{query\}}''.

4. The new summary should have the format with **Summary** and **Insights**.

5. The new summary should take into account the changes of my long-term interest over time.

6. My activities related to this time range are cataloged in the following ``User Activities'' section, with each separated by a newline.

7. Limit the summary to no more than \textbf{\{max\_words\}} words.
\newline
\newline
----- Previous Summary -----
\newline
\textbf{\{previous\_summary\}}
\newline
\newline
----- Previous Question Answers -----
\newline
\textbf{\{previous\_question\_answer\_pairs\}}
\newline
\newline
----- User Activities -----
\newline
\textbf{\{user\_activities\}}
}
\end{mdframed}

Here we define:

\textbf{\{time\_range\}:} Time range of the segment user activities.
    
\textbf{\{query\}:} Query intention of the summarization. For example, "Summarize my long-term movie watching preference".
    
\textbf{\{max\_words\}:} Max number of words for the summary.

\textbf{\{previous\_summary\}:} Previous generated summary.

\textbf{\{previous\_question\_answer\_pairs\}:} Previous list of generated QA pairs.

\textbf{\{user\_activities\}:} List of user activities of the segment.

\subsection{LLMs Prompt Template for Query Consistency} \label{appendix:query_consistency_prompt}

\begin{mdframed}
\small{
----- Instructions -----

Evaluate the relevance of a summary under the following ``Summary'' section to a query under the following ``Query'' section.

Return ``\textit{consistent}'' if the summary aligns with the query, and ``\textit{inconsistent}'' if the summary is unrelated to the query.

The response is only single word ``\textit{consistent}'' or ``\textit{inconsistent}'' without any explanation.
\newline
\newline
----- Summary -----
\newline
\textbf{\{summary\}}
\newline
\newline
----- Query -----
\newline
\textbf{\{query\}}
}
\end{mdframed}

Here we define:

\textbf{\{summary\}:} Provided summary.

\textbf{\{query\}:} Query intention of the summarization.

\subsection{LLMs Prompt Template for Question Generation}
\label{appendix:qg_prompt}

\begin{mdframed}
\small{
----- Instructions -----

Given the below ``KG Entities'' and ``Summary'', adhere to the instructions below to create ``Question-Answer Pairs'':

1. Each pair must be related to a specific KG entity.

2. The answer must be the KG entity itself.

3. Formulate questions that are directly relevant to the KG entity within the context of the summary.

4. Avoid creating questions that are open-ended.

5. Use the following format for your response, as shown under the ``Question-Answer Pairs'' of the ``Example'' section below: [Question\#1: \textit{"Question"}, Answer\#1: \textit{"Answer"}].

Following the above steps and an example under the following ``Example'' section, create ``Question-Answer Pairs'' based on the task under the ``Task'' section.
\newline
\newline
----- Example -----
\newline
``KG Entities'':
\newline
hiking
\newline
pop music
\newline
\newline
``Summary'':

**Summary:**

The user demonstrates a robust long-term interest in outdoor and musical activities. Specifically, they are drawn to hiking and pop music.

**Insights:**

* Sports Recreation and Fitness: The user has a sustained interest in hiking, engaging regularly in this activity, which indicates a preference for exploring nature and challenging terrains.

* Entertainment Media and Arts: The user enjoys pop music, known for its wide appeal and catchy melodies, reflecting a consistent interest in this genre.
\newline
\newline
``Question-Answer Pairs'':

[Question\#1: ``What outdoor activity is the user mainly interested in according to their searches and discussions?'', Answer\#1: ``hiking'']

[Question\#2: ``What genre of music does the user prefer, known for its wide appeal and catchy melodies?'', Answer\#2: ``pop music'']
\newline
\newline
---Task---
\newline
``KG Entities'':
\newline
\textbf{\{kg\_entities\}}
\newline
\newline
``Summary'':
\newline
\textbf{\{summary\}}
}
\end{mdframed}

Here we define:

\textbf{\{kg\_entities\}:} List of KG entities extracted from the summary.

\textbf{\{summary\}:} Provided summary.

\subsection{LLMs Prompt Template for Question Answering}
\label{appendix:qa_prompt}

\begin{mdframed}
\small{
---Instructions---

Given the below ``Question-Answer Pairs'' and my ``User Activities'', judge if each question-answer pair is consistent with my activities.
Adhere to the instructions below.

1. Each ``Judgement'' is composed of ``Status'' and ``ReferenceAnswer'' like the following format: [Status\#1: \textit{"Status"}, ReferenceAnswer\#1: \textit{"ReferenceAnswer"}].

2. The ``Status'' should be labeled as ``\textit{consistent}'' or '\textit{inconsistent}''.

3. A ``\textit{consistent}'' status means the question-answer pair aligns with or does not contradict the information provided in my activities. The ``ReferenceAnswer'' should be ``none''.

4. An ``\textit{inconsistent}'' status means the question-answer pair conflicts directly with or is contradicted by the information provided in my activities. The ``ReferenceAnswer'' should be a new answer of the question based on my activities.

5. Use the following format for your response, as shown under the ``Judgements'' of the following ``Example'' section below like [Status\#2: \textit{"Status"}, ReferenceAnswer\#2: \textit{"ReferenceAnswer"}].

6. Match each judgement to its corresponding question-answer pair by their sequence, such as [Status\#2: \textit{"Status"}, ReferenceAnswer\#2: \textit{"ReferenceAnswer"}] pertains to 
[Question\#2: \textit{"Question"}, Answer\#2: \textit{"Answer"}].

Following the above steps and an example under the following ``Example'' section, create judgements based on the task under the ``Task'' section.
\newline
\newline
----- Example -----
\newline
``Question-Answer Pairs'':

[Question\#1: ``What outdoor activity is the user mainly interested in according to their searches and discussions?'', Answer\#1: ``hiking'']

[Question\#2: ``What genre of music does the user prefer, known for its wide appeal and catchy melodies?'', Answer\#2: ``rock music'']
\newline
\newline
``User Activities'':
\newline
searched ``Pop music trends in the 2020s'' around Sat 05/15/2004 4PM \newline
searched ``Best coffee brewing methods for home'' around Fri 06/11/2004 6PM \newline
searched ``How to prepare for a multi-day hiking trip'' around Wed 07/21/2004 7PM \newline
searched ``The evolution of electronic elements in pop music'' around Sun 04/28/2004 5PM \newline
searched ``Hiking trails with the best views in the U.S.'' around Mon 04/29/2004 1PM \newline
\newline
``Judgements'':
\newline
[Status\#1: ``\textit{consistent}'', ReferenceAnswer\#1: ``none'']
\newline
[Status\#2: ``\textit{inconsistent}'', ReferenceAnswer\#2: ``pop music'']
\newline
\newline
----- Task -----
\newline
``Question-Answer Pairs'':
\newline
\textbf{\{question\_answer\_pairs\}}
\newline
\newline
``User Activities'':
\newline
\textbf{\{user\_activities\}}
}
\end{mdframed}

Here we define:

\textbf{\{question\_answer\_pairs\}:} List of generated QA pairs related to the KG entities.

\textbf{\{user\_activities\}:} User activities which are used to generate the summary.

\subsection{LLMs Prompt Template for Combining Time Segments}
\label{appendix:combine_segments}

\begin{mdframed}
\small{
---Instructions---

Combine all the time segment summaries under the ``Time Segment Summaries'' section. Adhere to the instructions below.

1. The combined summary should offer insights relevant to the query: ``\textbf{\{query\}}''.

2. Format the combined summary with sections labeled **Summary** and **Insights**.

3. Focus the combined summary on my recent interest preferences (recent time segments).

4. Limit the combined summary to no more than \textbf{\{max\_words\}} words.
\newline
\newline
----- Time Segment Summaries -----
\newline
Summary of Time Segment ``\textbf{\{time\_range\}}'':
\textbf{\{segment\_summary\}}

...
}
\end{mdframed}

Here we define:

\textbf{\{query\}:} Query intention of the summarization.

\textbf{\{max\_words\}:} Max number of words for the summary.

\textbf{\{time\_range\}:} Time range of a segment.

\textbf{\{segment\_summary\}:} Summary of a time segment.

\section{Analysis of Benchmark Metrics on Popular Models}
\label{appendix:benchmark_comparisons_models}

\begin{figure}[t]
    \centering
    \begin{subfigure}[b]{0.46\textwidth}
        \centering
        \includegraphics[width=\textwidth]{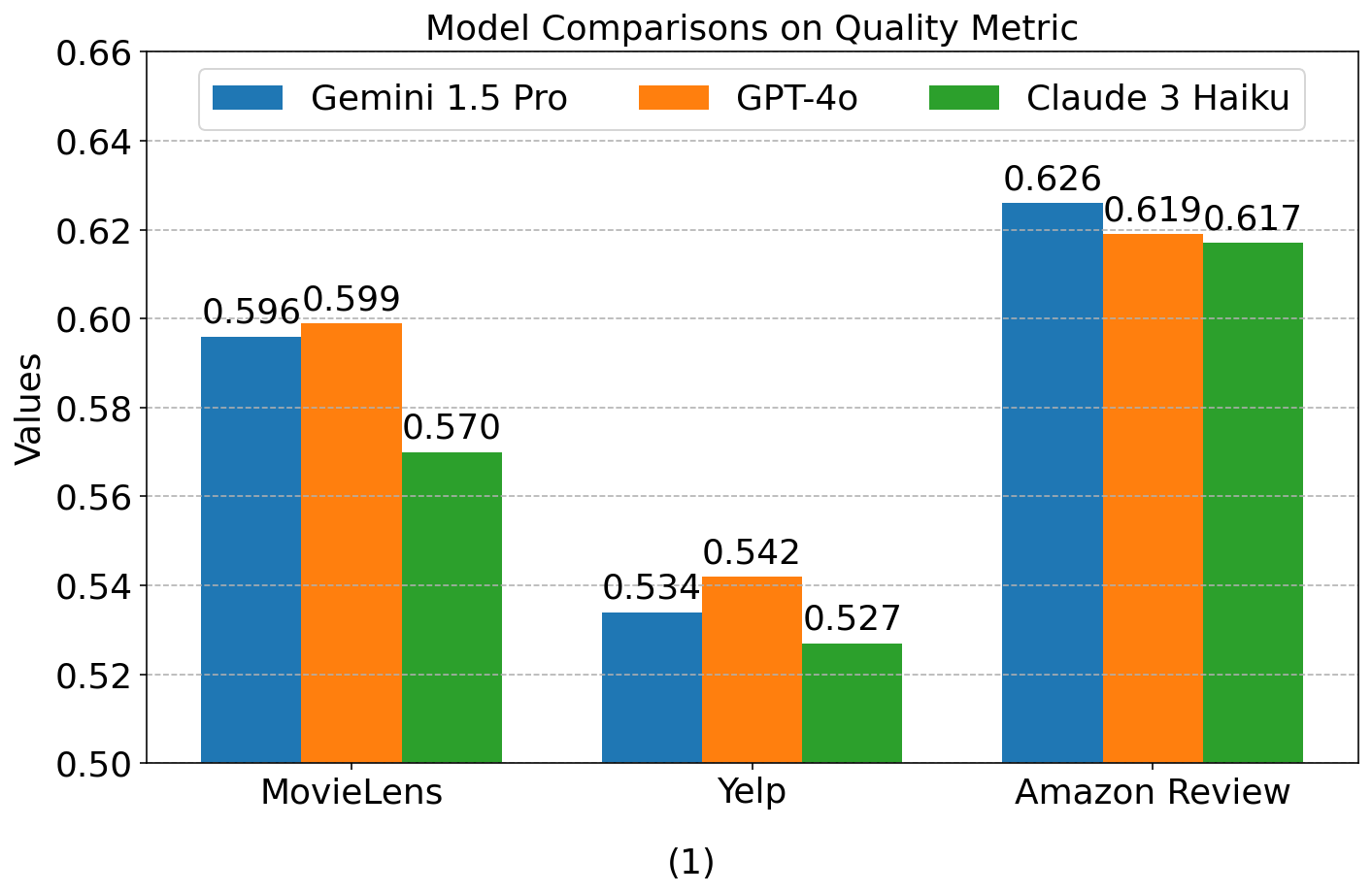}
    \end{subfigure}
    \hfill
    \begin{subfigure}[b]{0.46\textwidth}
        \centering
        \includegraphics[width=\textwidth]{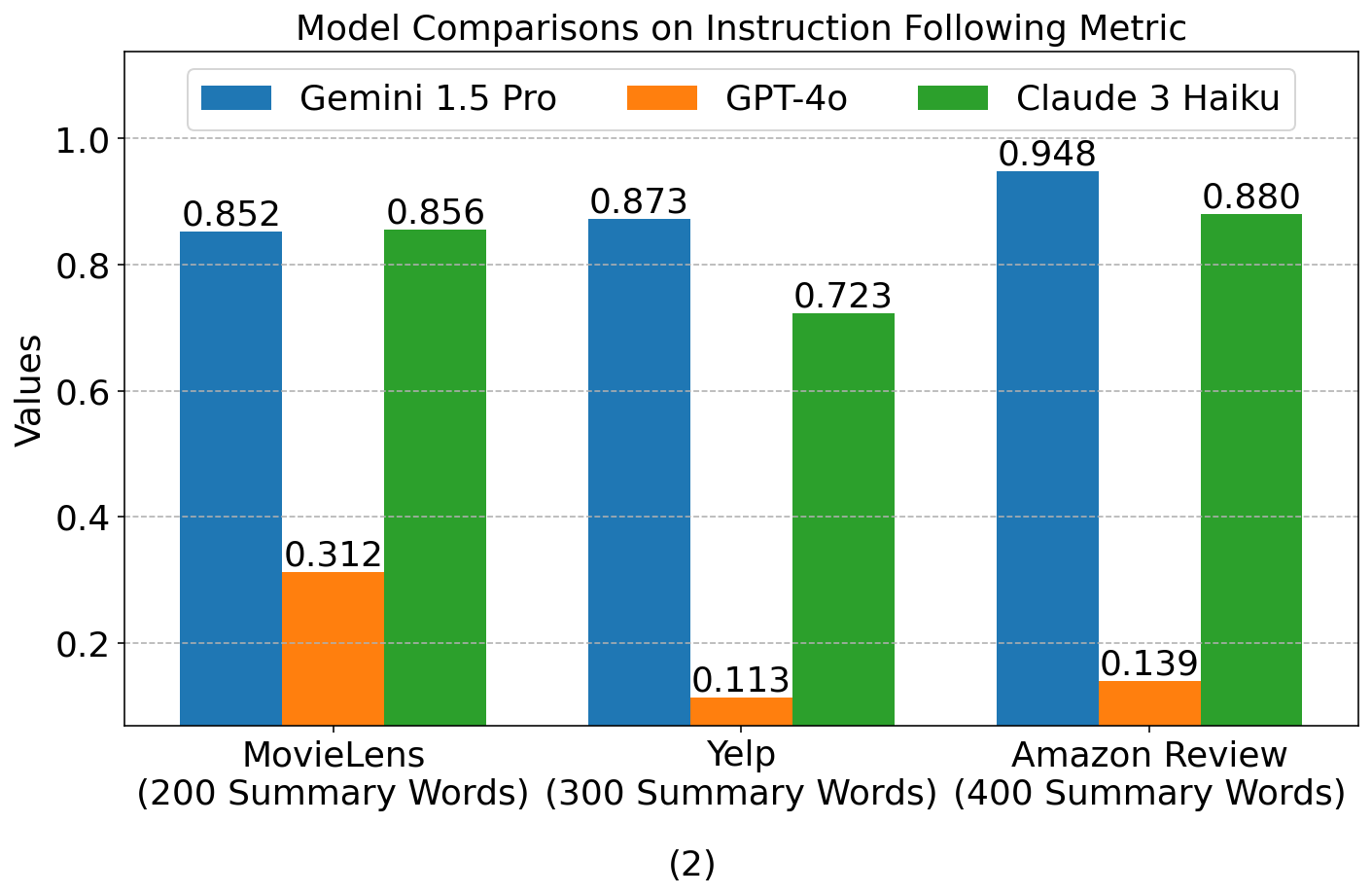}
    \end{subfigure}
    \hfill
    \begin{subfigure}[b]{0.46\textwidth}
        \centering
        \includegraphics[width=\textwidth]{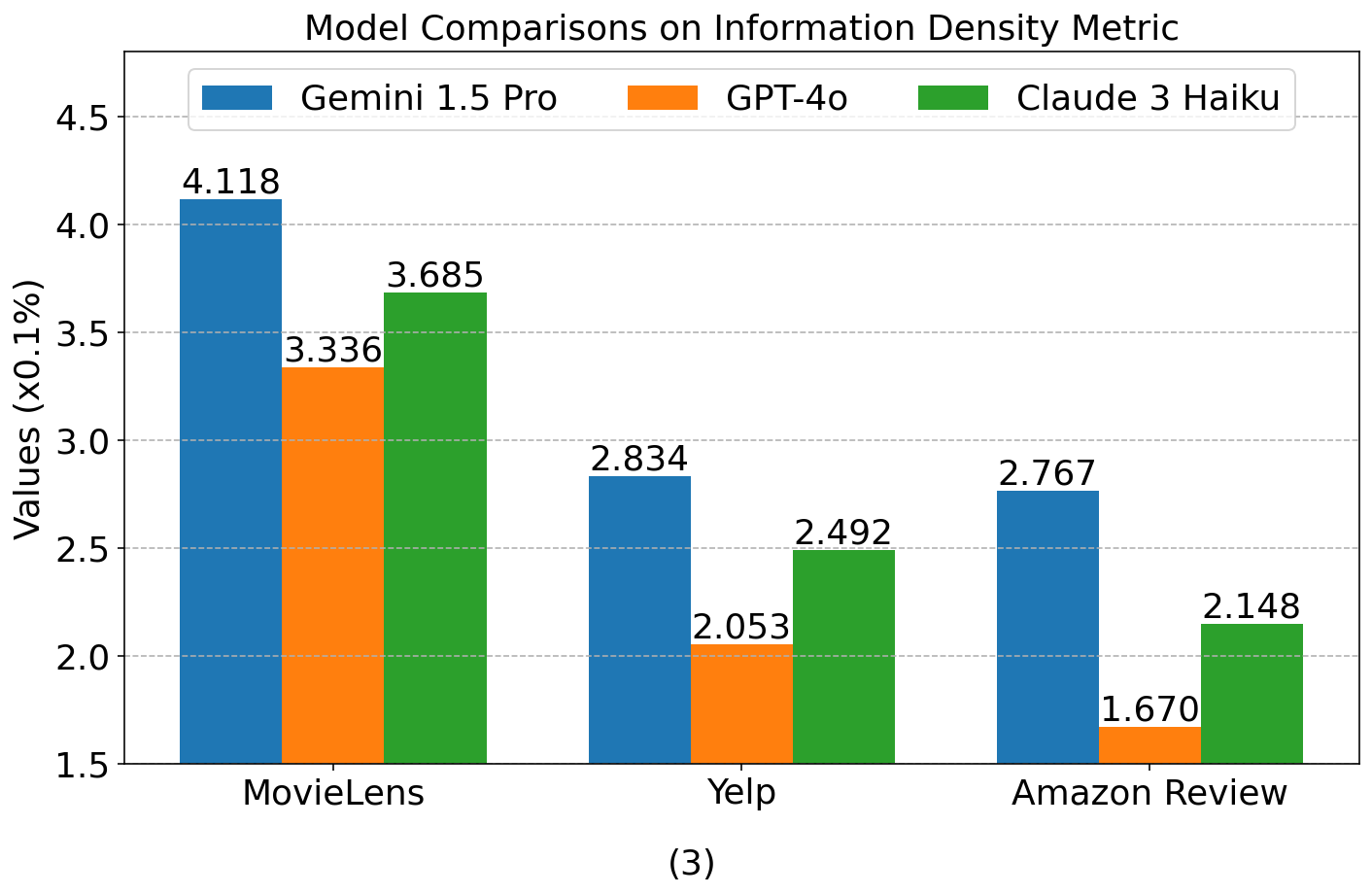}
    \end{subfigure}
    \caption{Comparison of different models (Gemini 1.5 Pro, GPT-4o, and Claude 3 Haiku) on various metrics across the MovieLens, Yelp, and Amazon Review datasets: (1) Quality Metric, (2) Instruction Following Metric, and (3) Information Density Metric.}
    \label{figure:model_comparison}
\end{figure}

We assessed three popular models for summarizing user activities (refer to Appendix~\ref{appendix:single_step_summarization_prompt} for the prompt details) using the \UserSumBench benchmark metrics. The models evaluated were Gemini 1.5 Pro \cite{reid2024gemini}, GPT-4o \cite{achiam2023gpt}, and Claude 3 Haiku \cite{anthropic2023claude3}. The summaries generated by these models were subsequently used for predicting future activities with the Gemini 1.5 Flash model \cite{reid2024gemini}. The results are presented in Figures~\ref{figure:model_comparison}.

\subsection*{Quality Metric}

Figure~\ref{figure:model_comparison} (1) shows the quality metric for the three models across the MovieLens, Yelp, and Amazon Review datasets. Gemini 1.5 Pro and GPT-4o exhibit similar performance in the three datasets. In the MovieLens and Yelp datasets, GPT-4o achieves the highest quality metric, surpassing both Gemini 1.5 Pro and Claude 3 Haiku. Claude 3 Haiku consistently shows slightly lower performance across all datasets but remains competitive, particularly in the Amazon Review dataset where it closely follows the other two models.

\subsection*{Instruction Following Metric}

Figure~\ref{figure:model_comparison} (2) compares the models based on the instruction-following metric. Gemini 1.5 Pro performs very well in all datasets, particularly in the Yelp and Amazon Review dataset, where it significantly outperforms the other models. Claude 3 Haiku also performs well, especially in the MovieLens dataset. GPT-4o, however, shows significantly lower performance in following instructions, particularly in the Yelp and Amazon Review datasets.

The observed poorer performance of GPT-4o on the instruction-following metric, particularly regarding summary word limits, could be attributed to its inherent design and optimization focus. Unlike some models that may be explicitly tuned for concise responses or instruction adherence, GPT-4o might prioritize generating detailed, comprehensive content, even at the expense of brevity. This tendency to produce more elaborate summaries could explain its weaker performance in adhering to strict word limits.

While GPT-4o's tendency to generate more detailed responses may hinder its ability to strictly follow word limits, it may simultaneously contribute to its higher quality in predicting future activities. The additional detail and context provided in longer summaries could lead to richer representations of user behaviors and preferences, enhancing predictive accuracy on tasks such as those in the MovieLens and Yelp datasets as shown in Figure~\ref{figure:model_comparison} (1).

In summary, GPT-4o's architecture or training objectives may emphasize content richness over strict instruction adherence, which, while beneficial for some tasks, leads to challenges in settings where brevity and instruction following are critical.

\subsection*{Information Density Metric}

Figure~\ref{figure:model_comparison} (3) presents the information density metric. Gemini 1.5 Pro demonstrates the highest performance across all datasets. Claude 3 Haiku also performs well but is consistently outperformed by Gemini 1.5 Pro. GPT-4o shows the lowest performance, particularly in the Yelp and Amazon Review datasets, where the difference is most pronounced.

\subsection*{Overall Analysis}

Overall, Gemini 1.5 Pro exhibits the best performance across all benchmark metrics and datasets, particularly excelling in instruction following and information density metrics. GPT-4o, while competitive in the quality metric, falls behind significantly in instruction following and information density metrics. Claude 3 Haiku shows consistent performance across the board but does not surpass Gemini 1.5 Pro in most of the metrics.

These results indicate that Gemini 1.5 Pro is the most effective model among the three for generating high-quality, instruction-following, and information-dense summaries. This superior performance makes it a preferable choice for applications requiring detailed and accurate summarization capabilities.

\end{document}